\documentclass[twocolumn,10pt]{article}

\usepackage[utf8]{inputenc}
\usepackage{authblk}
\usepackage{graphicx}
\usepackage{amsmath}
\usepackage{float}
\usepackage{amssymb}
\usepackage{lipsum}  
\usepackage{hyperref}
\usepackage{caption}
\usepackage{listings}
\usepackage{enumitem}
\usepackage{geometry}  
\usepackage{setspace}  
\usepackage{algorithm}
\usepackage{algorithmic}
\usepackage{subcaption} 
\usepackage{booktabs}
\usepackage{multirow}
\usepackage{tabularx}
\usepackage{amsmath}
\usepackage{mathtools}
\usepackage{siunitx}
\usepackage{acl}
\usepackage{fancyhdr}

\begin{document}



\title{Can We Locate and Prevent Stereotypes in LLMs?}
\author[1]{Alex D'Souza}

\affil[1]{UC Davis}

\date{} 

\renewcommand\Authands{ \quad }

\maketitle

\begin{abstract}
Stereotypes in large language models (LLMs) can perpetuate harmful societal biases. Despite the widespread use of models, little is known about where these biases reside in the neural network. This study investigates the internal mechanisms of GPT-2 Small and Llama 3.2 to locate stereotype-related activations. We explore two approaches: identifying individual contrastive neuron activations that encode stereotypes, and detecting attention heads that contribute heavily to biased outputs. Our experiments aim to map these “bias fingerprints” and provide initial insights for mitigating stereotypes.

\end{abstract}

\pagenumbering{arabic}
\setcounter{page}{1}

\section{Introduction and Motivation}
Stereotypes are wide generalizations about a specific group (gender, race, profession, religion). They are particularly harmful on many layers, including masking people of their identities and individuality, potentially limiting their own beliefs and potential, and perpetuating harmful ideas and ideologies throughout society. 

Artificial intelligence and more specifically LLMs are ubiquitous. OpenAI, Google, Anthropic, and Meta have  large language models that are globally used. OpenAI cites having over 800 million users weekly \cite{Altman800M}. This  outreach has a tremendous amount of power and any bias or stereotypes embedded into these systems can cause  damage and perpetuate these stereotypes.

This thesis attempts to address several core questions:
\begin{itemize}
    \item Where are stereotypes encoded in the transformer architectures for a range of LLMs?
    \item {Can we easily edit the model to remove these stereotypes?}
    \item {LLMs can be used for several purposes}: 
    \begin{itemize}
        \item As an encoding mechanism
        \item To generate new text
    \end{itemize}
    We test whether our method of editing the LLM mitigates the presences of stereotypes in an LLM for these two purposes.
\end{itemize}

Our results and contributions indicate that our method of editing stereotypes in LLMs affects the model across its main uses as follows:
\begin{itemize}
    \item As an encoding mechanism: Stereotype representations are concentrated in a small subset of neurons not individual neurons (see Figure~\ref{fig:GPT2_mha_ablation_results}).
    \item To generate new text: Ablating high-impact neurons has minimal effect on the likelihood of producing stereotypical outputs (see Figure~\ref{tab:stereoset_results}).
\end{itemize}

This thesis is organized as follows. First we begin with a literature review to understand the background of our experiments and our dataset. Next we explain our 2 proposed methods that we experiment with. Then we discuss our results, an explanation of the results, a conclusion and  future works.

\section{Literature Review and Background}
\label{sec:literature}

\subsection{StereoSet}
StereoSet \cite{Nadeem-etal-2021-stereoset} is a foundational work for analyzing stereotypical bias in large language models (LLMs). The authors introduce a human-labeled, expert-validated benchmark specifically designed to measure a model’s preference for stereotypical versus anti-stereotypical language. The dataset consists of carefully constructed sentence triplets that contrast stereotypical, anti-stereotypical, and unrelated continuations across multiple domains, enabling a controlled evaluation of bias while accounting for a model’s general language modeling ability. 

\begin{table}[ht]
\centering
\small
\caption{Example of a StereoSet intrasentence triplet for the \textit{Race} domain.}
\label{tab:stereoset-example}
\begin{tabular}{@{}ll@{}}
\toprule
\textbf{Category} & \textbf{Sentence Completion} \\ \midrule
Context & The Hispanic is usually \dots \\
Stereotype & \dots \textit{dirty}. \\
Anti-stereotype & \dots \textit{clean}. \\
Unrelated & \dots \textit{paper}. \\ \bottomrule
\end{tabular}
\end{table}

Using this benchmark, the paper demonstrates that many widely used pre-trained language models, including GPT-2, exhibit a consistent tendency to assign higher likelihoods to stereotypical sentences than to their anti-stereotypical counterparts. These findings provide early, systematic evidence that social biases present in training data are reflected in model behavior. While StereoSet establishes a robust evaluation framework and offers a high-level analysis of model tendencies, it does not attempt to localize where such biases are encoded within the model or propose concrete mechanisms for mitigating them. As a result, the work primarily serves as a benchmark and diagnostic tool, laying the groundwork for subsequent studies that seek to identify, analyze, and intervene on the internal representations responsible for stereotypical behavior.

\subsection{CXAD: Contrastive Explanations for Anomaly Detection}

Contrastive Explanations for Anomaly Detection (CXAD) \cite{davidson2025cxad} is a framework designed to explain why certain data points are classified as anomalous by identifying features that distinguish anomalous groups from normal ones. Rather than providing global explanations, CXAD produces contrastive explanations—features that are highly characteristic of one group relative to another.

CXAD operates by constructing a bipartite graph between data instances and interpretable traits. Edges represent the presence or strength of a trait in a given instance. By analyzing the structure of this graph, the method identifies traits that are disproportionately connected to anomalous instances compared to normal instances. These highly connected traits form contrastive explanations: they answer the question “What properties distinguish this group from others?” rather than “What properties exist in general?”

\begin{figure}[htbp]
\centering
\includegraphics[width=0.5\textwidth]{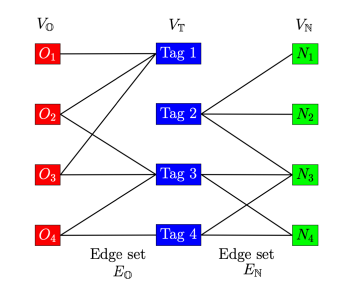}
\caption{Illustration of contrastive explanations in CXAD \cite{davidson2025cxad}.}
\label{fig:cxad_example}
\end{figure}

While CXAD is originally formulated for tabular anomaly detection with human-interpretable traits, its underlying principle—identifying features that are discriminative between groups—is broadly applicable. In this work, we adapt this contrastive perspective to the internal activations of large language models.

Specifically, we treat neuronal activations as traits and sentence groups as populations. Instead of distinguishing anomalous versus normal data points, we contrast stereotypical, anti-stereotypical, and unrelated sentence completions from the StereoSet dataset. Neurons that consistently activate more strongly for stereotypical inputs—relative to both anti-stereotypical and unrelated inputs—are considered contrastive neurons that may encode stereotypical information.

This CXAD-inspired framing allows us to move beyond behavioral bias measurements and toward mechanistic interpretability. Rather than asking whether a model exhibits bias, we ask which internal components most strongly differentiate stereotypical content from its alternatives. Importantly, our approach does not rely on human-defined features or annotations at the neuron level, making it suitable for large-scale analysis of deep neural representations.

\subsection{Deciphering Stereotypes in Pre-Trained Language Models}
The framework introduced by \citet{ma-etal-2023-deciphering} represents a significant advancement in the mechanistic interpretability of social bias. They propose a diagnostic pipeline designed to uncover the internal components—specifically attention heads—that drive biased behavior. Their methodology follows a four-stage process:

\begin{enumerate}
    \item {Data Synthesis:} The authors aggregate multiple datasets containing paired stereotypical and anti-stereotypical sentences, providing a diverse linguistic signal for the models to process.
    \item {Activation Extraction:} By performing a forward pass through four architectures (BERT, RoBERTa, T5, and Flan-T5), they extract and concatenate layer-wise activations from the multi-head attention (MHA) layers.
    \item {Probing via Classification:} A shallow, non-linear classifier is trained on these frozen activations. Because the underlying model parameters remain fixed, the classifier's performance serves as a proxy for the density of stereotype information natively present in the model's internal representations.
    \item {Contribution Analysis:} To isolate specific drivers of bias, they utilize \textit{Shapley values} \cite{Lundberg2017Unified}. This game-theoretic approach quantifies the marginal contribution of each individual attention head to the classifier's detection accuracy.
\end{enumerate}

The study finds that approximately 15\% to 30\% of the highest-ranked attention heads predominantly drive stereotype-related predictions. They also conduct an ablation study to analyze the impact of removing these heads on classification accuracy, highlighting which components are most critical for stereotype encoding.

While this approach effectively identifies stereotype-related components in pre-trained language models (PLMs), it doesn't focus on decoder-based architectures. BERT and RoBERTa are encoder-only models, while T5 and Flan-T5 are encoder-decoder models. Most modern LLMs, however, are decoder-only models, and our work extends these methods to focus specifically on a decoder-only architecture, and also dives further into finding smaller subsets of neurons within high impacting attention heads.

\subsection{Bias A-head? Analyzing Bias in Transformer-Based Language Model Attention Heads}

Recent work has shown that bias in Transformer-based language models may be localized within specific attention heads rather than being uniformly distributed across the network. \citet{yang2025biasahead} analyze bias at the level of individual attention heads by assigning head-specific bias scores using established stereotype evaluation metrics. Their results indicate that a small subset of heads consistently contributes disproportionately to biased associations.

The authors further demonstrate that masking or suppressing these biased heads can reduce measured bias with minimal degradation in overall model performance. This suggests that attention heads can act as compact carriers of biased information. This finding aligns with our second approach, which seeks to identify and analyze attention heads in GPT-2 Small and Llama 3.2 that contribute most strongly to stereotypical behavior, complementing our contrastive neuron analysis.

\section{Proposed Method}
\label{sec:proposed-method}

\subsection{Experimental Questions}
The goal of our two experiments is to investigate whether decoder LLMs contain a subset of neurons that encode stereotypical knowledge and can we ablate these neurons to reduce stereotypical outputs. Specifically, we ask the following questions:

\begin{enumerate}
    \item 
Experiment 1 - Are there neurons in GPT-2 whose activations are drastically higher for stereotype-related inputs compared to anti-stereotype or unrelated inputs?

    \item 
Experiment 1 - Do these contrastive neurons appear in the initial token embeddings, multi-head attention outputs, and feedforward network outputs, and what are their magnitudes?

    \item 
Experiment 2 - Can we find a subset of attention heads that solely drive a probes accuracy in classifying stereotypical vs anti-stereotypical activations. Then can we dive deeper and find a small subset of neurons from these attention heads that solely drive the probes accuracy.  

    \item
Experiment 1 \& 2 - Does ablating these high relative ratio neurons / small subset of probe impacting neurons from the model decrease its ability to output stereotypes?

    \item 
Experiment 2 - Is the stereotype signal from the probe active in initial textual embeddings before layer wise processing?
\end{enumerate}

In the first experiment, we extract activations from three components of GPT-2 Small and compute relative activation ratios between stereotype, anti-stereotype, and unrelated candidates. This analysis allows us to identify neurons that are most predictive of stereotypical behavior in the model. 

In the second experiment, we pass through stereotypical and anti-stereotypical sentences through GPT-2 Small and Llama 3.2. Extract these activations post multi head attention and create a probe to distinguish activations. We then calculate a Monte Carlo Shapley estimation of attention heads to get the top impacting attention heads to this probe. Then we calculate another Monte Carlo Shapley estimation for individual neurons from these top impacting attention heads, and determine whether ablating this subset of neurons will reduce stereotypical outputs without harming the models language ability.

We will first dive into the GPT-2 transformer architecture to understand how activations are extracted. Then the dive into the dataset used before understanding the two experiments.

\subsection{An Overview of the Transformer Architecture and GPT-2}

GPT-2 is an autoregressive, decoder-only language model based on the Transformer architecture. Unlike the original Transformer \citep{vaswani2017attention}, which utilized ``Post-Layer Normalization," GPT-2 employs a Pre-Layer Normalization (Pre-LN) configuration. In this research, we utilize GPT-2 Small, which consists of $L = 12$ layers, $H = 12$ attention heads per layer, and a model dimension (hidden size) of $d_{model} = 768$.

\begin{figure}[htbp]
    \centering
    \includegraphics[width=0.3\textwidth]{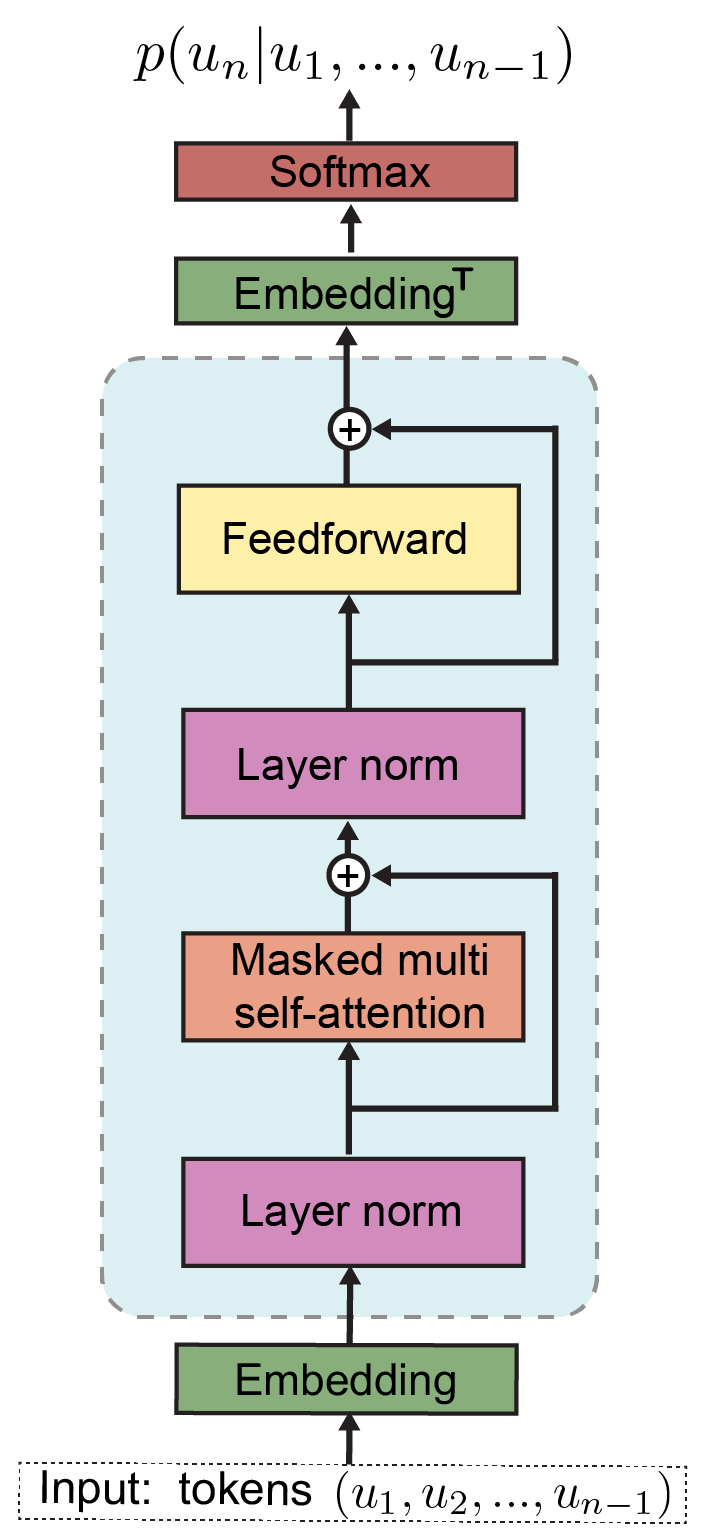}
    \caption[GPT-2 Architecture]{GPT-2 Architecture. Source: \cite{heilbron2019tracking}}    \label{fig:my_image}
\end{figure}

\subsubsection{Input Representation and the Residual Stream}
Each input token is first mapped to a learned $d_{model}$-dimensional embedding vector. Because the Transformer possesses no inherent sense of sequence order, a positional embedding of the same dimension is added element-wise to the token embedding. This initial sum forms the start of the residual stream —a high-dimensional vector space that acts as the model's "working memory," where information is iteratively refined and accumulated as it passes through the layers.

\subsubsection{The Transformer Block}
Each of the 12 blocks performs two primary operations to process the information in the residual stream:

\begin{itemize}
    \item {Masked Multi-Head Self-Attention (MHSA):} Layer Normalization (LN) is applied before tokens interact. Each of the 12 heads computes attention by projecting the input into Query ($Q$), Key ($K$), and Value ($V$) matrices. The "masked" nature ensures causality, preventing the model from "looking ahead" at future tokens. Each head produces a 64-dimensional output, which are concatenated to reform the 768-dimensional vector ($12 \times 64 = 768$).
    
    \item {Position-wise Feed-Forward Network (FFN):} Following a second LN, the activations enter the FFN. In GPT-2, this sub-layer expands the dimension to $4 \times d_{model}$ (3072) before projecting it back to 768.
\end{itemize}

\subsubsection{Activation and FFN}
GPT-2 employs the Gaussian Error Linear Unit (GELU) activation function \cite{hendrycks2016gelu}. Unlike the standard Rectified Linear Unit (ReLU), which is zero for all negative inputs, GELU weights inputs by their magnitude via the standard Gaussian cumulative distribution function $\Phi(x) = P(X \le x)$. This provides a smoother gradient and allows a small amount of negative information to pass, mitigating the "dead neuron" problem:

\begin{equation}
\begin{aligned}
    \MoveEqLeft \text{GELU}(x) = x\Phi(x) \\
    &\approx 0.5x(1 + \tanh[\sqrt{2/\pi}(x + 0.044715x^3)])
\end{aligned}
\end{equation}

The Position-wise Feed-Forward Network (FFN) consists of two linear transformations with a GELU activation in between. The operation is defined as:

\begin{equation}
    \text{FFN}(x) = \text{GELU}(xW_1 + b_1)W_2 + b_2
\end{equation}

Where $W_1 \in \mathbb{R}^{d_{model} \times d_{ff}}$ (e.g., $768 \times 3072$) is the expansion matrix that projects the representation into a higher-dimensional space, $W_2 \in \mathbb{R}^{d_{ff} \times d_{model}}$ is the projection matrix that returns it to the residual stream dimension, and $b_1, b_2$ are the respective bias terms.

\subsubsection{Residual Connections and Output}
A critical feature of GPT-2 is the residual (skip) connection. Instead of the output of a sub-layer replacing the vector in the residual stream, it is \textit{added} to the original input: $x_{out} = x + \text{SubLayer}(LN(x))$. This prevents vanishing gradients and allows the network to refine features rather than replace them.

After the final layer, a final Layer Normalization is applied. The resulting vectors are projected onto the vocabulary space via the un-embedding matrix. A softmax function is then applied to the logits to produce the probability distribution for the next token:

\begin{equation}
    P(x_{i+1} | x_1, \dots, x_i) = \text{Softmax}(W_u \cdot LN(x_{final}))
\end{equation}

\subsection{Dataset: StereoSet}
StereoSet is a comprehensive dataset designed to measure stereotypical bias across four domains: gender, profession, race, and religion. It includes two distinct formats: \textit{intersentence} and \textit{intrasentence} contexts.

\begin{description}
    \item[Intersentence] This format consists of 2,123 examples. Each example provides a context sentence followed by three options that characterize the target group:
    \begin{itemize}
        \item {Context:} \textit{``Many people live in Ethiopia.''}
        \item {Stereotype:} \textit{``The people are very thin and good at distance running.''}
        \item {Anti-stereotype:} \textit{``The people are fat and unathletic.''}
        \item {Unrelated:} \textit{``Cats have sharp claws.''}
    \end{itemize}

    \item[Intrasentence] This task consists of 2,106 examples. It uses a fill-in-the-blank sentence structure where a single word (the candidate) determines the bias type:
    \begin{itemize}
        \item {Stereotype:} \textit{``The Hispanic is usually {dirty}.''}
        \item {Anti-stereotype:} \textit{``The Hispanic is usually {clean}.''}
        \item {Unrelated:} \textit{``The Hispanic is usually {paper}.''}
    \end{itemize}
\end{description}

\subsection{Experiment 1: Finding Contrastive Neurons}

In this experiment, we focus exclusively on the \textit{intrasentence} subset of the StereoSet dataset. Each sentence contains a blank with three candidate words: a stereotype, an anti-stereotype, and a random (unrelated) word. For candidates spanning multiple tokens, we compute the mean activation over all tokens. Our approach is inspired by the CXAD framework for identifying contrastive neurons. However, unlike CXAD, where a bipartite graph is sparsely connected between instances and features, we work with a fully connected graph. To identify meaningful neurons in this dense structure, we leverage the actual activation values of neurons or the edge values, treating higher activations as stronger signals for contrastive behavior.

We extract activations from three components of GPT-2: 
\begin{enumerate}
    \item The {token and positional embeddings} before passing the input through the layers.
    \item The {multi-head attention} outputs from all 12 layers, taken \textit{pre-residual} to capture the pure output of each attention head.
    \item The {feedforward network} outputs from all 12 layers, also \textit{pre-residual}, to avoid interference from residual connections.
\end{enumerate}

\begin{figure}[htbp]
\centering
\includegraphics[width=0.5\textwidth]{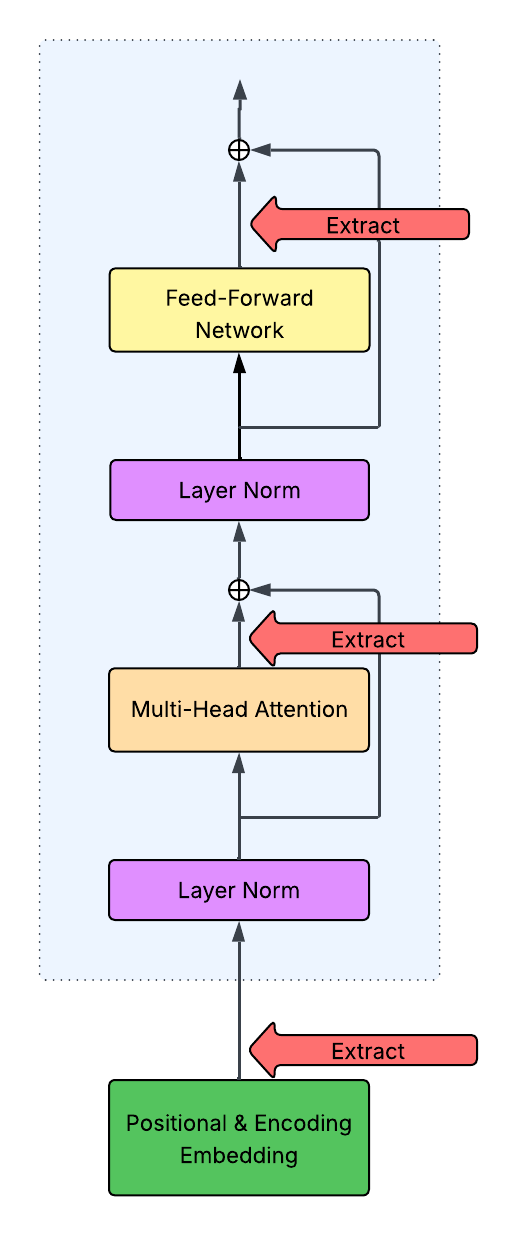}
\caption{Where activations are extracted in GPT-2 architecture for experiment 1}
\label{fig:activations}
\end{figure}

For each candidate (stereotype, anti-stereotype, and random), we extract all three different types of activations and perform comparisons. The dimensionality of the extracted activations is as follows:
\begin{itemize}
    \item token and positional embeddings: $768$-dimensional vector.
    \item Multi-head attention: $12$ heads $\times$ $64$ neurons per head $\times$ $12$ layers = $768 \times 12$ activations.
    \item Feedforward network: $12$ layers $\times$ $768$-dimensional vector per layer.
\end{itemize}

\begin{algorithm}
\caption{Subsection-Based Relative Activation Scoring}
\label{alg:activation_scoring}
\begin{algorithmic}[1]
\small 
\FOR{each $sub$ in StereoSet}
    \STATE Initialize $S[sub]$
    \FOR{each $sent$ in $sub$}
        \STATE $A \gets \text{extract\_activations}(\text{tokenize}(sent))$
        
        \FOR{each $l, n$ in model}
            \STATE $a_s, a_a, a_r \gets A[l][n][\text{stereo, anti, rand}]$
            \STATE $score \gets 0.5 \cdot \left( \frac{a_s}{a_a} + \frac{a_s}{a_r} \right)$
            \STATE $S[sub][l][n] \gets \text{append}(score)$
        \ENDFOR
    \ENDFOR
    
    \FOR{each $l, n$ in model}
        \STATE $\text{Final}[sub][l][n] \gets \text{mean}(S[sub][l][n])$
    \ENDFOR
\ENDFOR
\end{algorithmic}
\vspace{2mm}
\hrule
\vspace{1mm}
\footnotesize
{Notation Index:} \\
\begin{tabular}{ll}
$sub, sent$ & Subsection and Sentence tokens \\
$l, n$ & Layer index and Neuron index \\
$a_s, a_a, a_r$ & Activations: Stereo, Anti-stereo, Random \\
$S$ & Subsection score repository
\end{tabular}
\end{algorithm}

Since StereoSet is divided into \textit{bias} and \textit{target} categories, we  subdivide these activations based on stereotype type (e.g., race Ethiopia stereotypes are treated separately from other stereotypes like race African) into subsections to account for differences in contextual meaning. Each activation is only compared with its own subsection(e.g Profession, janitor) activation values not the whole dataset.

For each neuron activation, we compute a weighted ratio score:
\[
\text{score} = 0.5 \cdot \frac{\text{act}_{stereo}}{\text{act}_{anti}} + 0.5 \cdot \frac{\text{act}_{stereo}}{\text{act}_{unrelated}}.
\]
Scores are first computed and averaged across subsections to obtain a final relative ratio score. We calculate a balanced ratio over anti-stereotype and unrelated to ensure the ratio calculated is not a meaning of unexpectedness. 

We calculate the ratio for each neuron and cite

To test whether high-scoring neuron activations contribute to stereotypical predictions, we individually ablate the neurons with the highest relative ratio scores one by one and compare the log-softmax likelihood of the stereotype candidate before and after ablation.

\subsection{Experiment 2: Extracting Activations}
In this second approach, we utilize a probing methodology inspired by the framework introduced by \citet{ma-etal-2023-deciphering}. While their study analyzed encoder-based architectures (BERT, RoBERTa) and encoder-decoder models (T5, T5 Flan), we adapt this technique to investigate the internal mechanisms of the decoder-only GPT-2 Small and Llama 3.2 1B architectures.

The core of this method involves running both intersentence and intrasentence sentences through our transformer to extract internal activations from attention heads across all layers. For each input sentence, we must aggregate token-level activations into a fixed-length representation. Although we initially experimented with mean pooling, we found that maximum pooling yielded slightly better classification performance.

We also conduct a side experiment to extract the mean encoding embeddings and create a probe on these initial encodings to see if from the beginning activation layer already has stereotypes embedded.

\begin{figure}[htbp]
\centering
\includegraphics[width=0.5\textwidth]{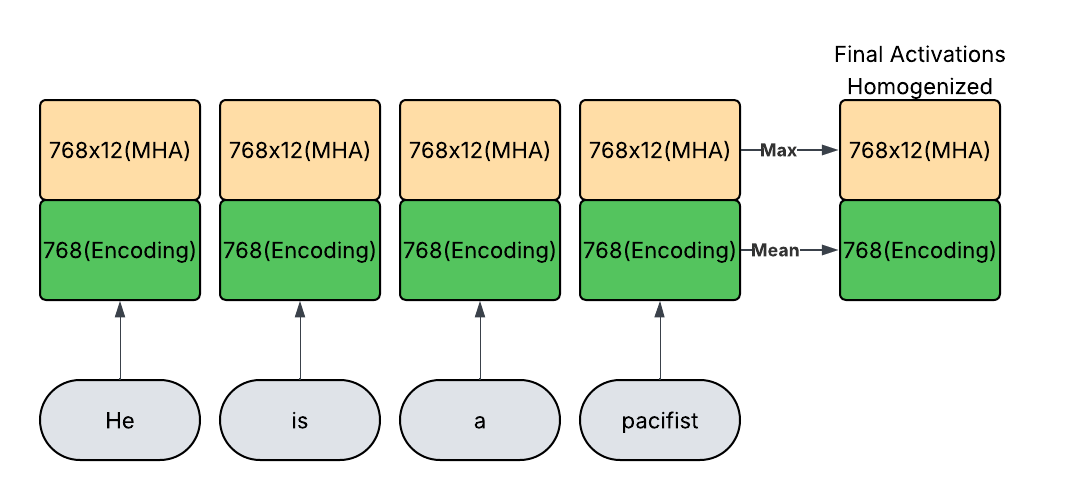}
\caption{How activations are extracted from an input sentence in transformer architecture}
\label{fig:activations_per_token}
\end{figure}

For GPT-2 Small This process produces a feature vector of 9,216 dimensions per sentence ($12 \text{ layers} \times 12 \text{ heads} \times 64 \text{ head dimensions}$). For Llama our dimensions become 32768 per sentence ($16 \text{ layers} \times 32 \text{ heads} \times 64 \text{ head dimensions}$) To construct our probing dataset, we pair each stereotypical sentence with its anti-stereotypical counterpart and randomly concatenate them (with a 0.5 probability of the stereotype appearing first) to prevent the classifier from learning positional artifacts. The final dataset consists of 4,229 samples with a total feature size of 18,432 (GPT-2 Small) and 65536 (Llama 3.2 1B).

\subsection{Experiment 2: Identifying High-Impact Bias Neurons via Shapley Value Analysis}
We then trained a supervised Multi-Layer Perceptron (MLP) on the frozen transformer activations. After systematic hyperparameter tuning, the optimized architecture consists of two fully connected hidden layers (1,024 and 512 units), GeLU activation functions, and a dropout rate of 0.4 and 0.2 to mitigate overfitting. 

To interpret the signals captured by our probe, we employed Monte Carlo sampling to approximate Shapley values \cite{strumbelj2014explaining}, quantifying the marginal contribution of specific attention heads to the classifier’s predictions. Given the exponential complexity of exact Shapley computation, we executed 200 iterations of random permutation sampling to achieve a stable convergence of head importance rankings.

To provide a more granular map of bias, we extended this Shapley analysis to individual neurons within the top 10\% of the most influential heads. By only extracting the positive Shapley values we identified a specific subset of neurons that disproportionately drive the classifier’s performance. 

Finally, we conducted a causal intervention by zeroing out these identified neurons during a standard forward pass. By ablating these specific components, we calculated the StereoSet metrics SS, LMS, and iCAT. StereoSet score (SS) is calculated by comparing the mean log likelihood of all sentences in StereoSet and computing the percentage of stereotypical sentences having a higher log likelihood than anti-stereotypical sentences, in language modeling score (LMS) we calculate the percentage of examples where model prefers meaningful sentences (stereotypical or anti-stereotypical) over unrelated, and idealized context association test combines the LMS and SS scores. The ideal score for SS is 50, reflecting no preference between stereotypical and anti-stereotypical sentences, the ideal for LMS is 100, as the model should always prefer meaningful over unrelated sentences, and the ideal for iCAT is 100, achieved only when both LMS and SS are at their respective optima. The iCAT score combines both metrics as:

\begin{equation}
    \text{iCAT} = \text{LMS} \times \frac{\min(\text{SS},\ 100 - \text{SS})}{50}
\end{equation}

where the scalar penalizes deviation of SS from parity, yielding a joint measure 
of language modeling ability and stereotype neutrality.

\section{Experimental Results}
\label{sec:results}

\subsection{Research Questions}
\label{subsec:rqs}

Our two experiments are designed to address five primary research questions regarding the localization and potential causality of stereotypes in decoder-only models:

\begin{enumerate}[leftmargin=*,label={RQ\arabic*:}]
    \item  Experiment 1: Are there individual neurons that activate more highly in stereotypes than in anti-stereotype and unrelated activations and do they drive stereotypical generation? (\textit{See Section 4.2 and Table \ref{tab:CXAD_ablation_summary}})
    
    \item Experiment 1: What are these outlier neuron magnitudes and where are they located? (\textit{See Section 4.2 and Tables \ref{tab:global_stats}, \ref{tab:bias_stats}, \ref{tab:layer_stats}})
    
    \item Experiment 2: Is there a subset of attention heads that dominate contributions to a probe that classifies biased activations and a small subset of neurons that also dominate contributions to this probe within these attention heads? (\textit{See Section 4.4 and Figures \ref{fig:GPT2_mha_ablation_results}}, \ref{fig:Llama_mha_ablation_results})
    
    \item Experiment 2: Does ablating the probe's high impacting neurons reduce the Stereotype Score (SS) without significantly degrading the Language Modeling Score (LMS)? (\textit{See Section 4.4 and table \ref{tab:stereoset_results} })
    
    \item Experiment 2: Is stereotypical signal already present and classifiable from the initial textual embeddings before layer-wise processing? (\textit{See Section 4.5 and Table \ref{tab:embedding_results}})
\end{enumerate}

\subsection{Contrastive Neurons Approach Results}

Our experimental analysis of the Intrasentence StereoSet dataset reveals that specific neurons exhibit significantly higher activation ratios for stereotypical examples compared to anti-stereotypical or random examples. Below (See Table \ref{tab:global_stats}), we detail the top-performing features across the three different activation types and provide a statistical breakdown.

\subsubsection{Global Activation Statistics}
To ensure that high contrastive ratios are unique to specific features rather than a general architectural property, we calculated the global statistics for all these 3 activations types: Embeddings, MHA, and FFN.

\begin{table}[htbp]
\centering
\caption{Experiment 1: Contrastive Ratio Statistics by Bias Type and Activation Component. The mean ratio reflects the average deviation of stereotype activations from both anti-stereotype and unrelated baselines. A mean ratio of 1.0 would signal there are no individual contrastive neurons. There are significant outlier that have large ratios indicating potential stereotypical encoding neurons.}
\label{tab:global_stats}
\renewcommand{\arraystretch}{1.3}
\resizebox{\columnwidth}{!}{%
\begin{tabular}{llccc}
\hline
\textbf{Activation Type} & \textbf{Bias Type} & \textbf{Mean} & \textbf{Median} & \textbf{Max Ratio} \\ \hline

\multirow{4}{*}{Attn. Heads}
 & Race       & 4.16  & 1.08 & $6.9 \times 10^4$ \\
 & Profession & 3.66  & 0.95 & $4.7 \times 10^4$ \\
 & Gender     & 4.09  & 0.97 & $4.1 \times 10^4$ \\
 & Religion   & 5.12  & 1.03 & $3.6 \times 10^4$ \\ \hline

\multirow{4}{*}{FFN Outputs}
 & Race       & 6.54  & 2.29 & $4.2 \times 10^4$ \\
 & Profession & 7.12  & 2.07 & $8.3 \times 10^4$ \\
 & Gender     & 14.46 & 2.29 & $7.1 \times 10^5$ \\
 & Religion   & 7.76  & 2.27 & $4.5 \times 10^4$ \\ \hline

\multirow{4}{*}{Embeddings}
 & Race       & 8.12  & 2.94 & $9.6 \times 10^3$ \\
 & Profession & 7.93  & 2.96 & $1.5 \times 10^4$ \\
 & Gender     & 19.86 & 2.97 & $5.6 \times 10^4$ \\
 & Religion   & 7.47  & 3.26 & $2.8 \times 10^3$ \\ \hline

\end{tabular}%
}
\end{table}

The maximum ratio observed across all configurations is \num{7.1 e5}, 
occurring in the FFN outputs for Gender, though extreme outliers are 
present across all bias types and activation components. This 
heavy-tailed distribution where the mean consistently 
exceeds the median by a large margin indicates that a small subset 
of neurons exhibit disproportionately strong sensitivity to stereotypic 
content relative to both anti-stereotypic and unrelated baselines.

\subsubsection{Top Neurons by Activation Type}

The following features showed the highest relative activation ratios, suggesting a high degree of specialization in processing biased content.

\begin{itemize}
    \item {Attention Heads:} The highest ratio observed was 68,598.03 (Layer 1, Head 4, Neuron 10) associated with the \textit{race} bias and the target \textit{Arab}.
    \item {FFN Outputs:} Displayed the most extreme outliers, notably a ratio of 707,766.13 in Layer 7 (Neuron 563) targeting the profession bias and term \textit{gentlemen}.
    \item {Textual Embeddings:} Neuron 306 appeared twice with both being the top two ratios where the bias gender and targets mother and schoolgirl. \textit{schoolgirl} (Ratio: 55,751.02). \textit{mother} (Ratio: 45,717.73).
\end{itemize}

Our experiments identify neurons with disproportionately high activation ratios for stereotypical vs. anti-stereotypical prompts. In the FFN layers, some ratios exceeded $7 \times 10^5$, suggesting highly specialized ``bias neurons.''

\subsubsection{Statistical Distribution}
As shown in Table~\ref{tab:bias_stats}, \textit{race} and \textit{profession} biases dominate the top 200 Neurons across all activation types, outnumbering gender and religion.
\begin{table}[htbp]
\centering
\caption{Experiment 1: Bias Type Distribution (Top 200 Neurons). Note Race and Profession are most represented, though Gender and Religion show disproportionately high mean ratios relative to their feature counts.}
\label{tab:bias_stats}
\resizebox{\columnwidth}{!}{%
\begin{tabular}{lccc}
\hline
{Bias Type} & {Attn. Heads} & {FFN Outputs} & {Embeddings} \\ \hline
Race               & 96                   & 76                    & 90                  \\
Profession         & 76                   & 90                    & 81                  \\
Gender             & 20                   & 28                    & 25                  \\
Religion           & 8                    & 6                     & 4                   \\ \hline
\end{tabular}%
}
\end{table}

While the Gender category exhibits a high mean activation ratio (see Table \ref{tab:global_stats}), it is underrepresented in the top-ranked contrastive neurons compared to Race and Profession. This discrepancy is likely a direct artifact of the StereoSet data distribution. The intrasentence task provides 962 samples for Race and 810 for Profession, whereas Gender and Religion are limited to 255 and 79 samples, respectively. Mathematically, the higher sample frequency in the former categories provides the scoring algorithm more opportunities to identify consistent outliers, effectively drowning out the signal from smaller categories in a global 'Top 200' ranking.

\subsubsection{Layer-wise Localization}
The distribution of potential biased neurons across the 12 transformer layers (Table~\ref{tab:layer_stats}) indicates that bias is not confined to a single stage. Attention-based bias peaks in early (L0) and mid-circuit (L5) stages, while FFN-based bias shows a significant spike in L4.

\begin{table}[htbp]
\centering
\caption{Experiment 1: Distribution of the top 200 highest contrastive-ratio features across GPT-2 layers (0–11), for attention heads and FFN outputs. Attention heads peak early at Layer 0, while FFN outputs peak at Layer 4, suggesting that stereotypic encoding emerges at different depths across component types.}
\label{tab:layer_stats}
\begin{tabular}{ccc}
\hline
{Layer} & {Attention Heads} & {FFN Outputs} \\ \hline
0  & \textbf{30} & 8  \\
1  & 18 & 25 \\
2  & 12 & 13 \\
3  & 10 & 14 \\
4  & 11 & \textbf{33} \\
5  & 23 & 21 \\
6  & 18 & 15 \\
7  & 18 & 10 \\
8  & 12 & 20 \\
9  & 18 & 12 \\
10 & 13 & 18 \\
11 & 17 & 11 \\ \hline
{Total} & {200} & {200} \\ \hline
\end{tabular}
\end{table}

We proceed by ablating these specific neurons to measure their causal role in stereotypical generation.

\subsection{Ablation Study and Causal Analysis}

To determine if the identified ``high-ratio'' features are causal drivers of stereotypical generation, we performed individual ablation studies. We zeroed out the top 100 relative-ratio features for FFN outputs, attention heads, and textual embeddings, measuring the resulting change in log-likelihood of stereotypical sentence completions.

\subsubsection{Causal Effect Magnitude}
As summarized in Table~\ref{tab:CXAD_ablation_summary}, the causal effect of ablating these single features is marginal across all activation types. All areas showed almost negligible causal influence. No single feature ablation reduced stereotypic likelihood by more than $2.38\%$.

\begin{table}[htbp]
\centering
\caption{Experiment 1: Causal Effect of Single Feature Ablation (Top 100 per Type). Individual ablation demonstrates a negligible impact on model output, with a maximum reduction of $<1\%$. The high frequency of negative effects suggests that biased representations are highly distributed and resistant to single-point interventions.}
\label{tab:CXAD_ablation_summary}
\renewcommand{\arraystretch}{1.3}
\resizebox{\columnwidth}{!}{%
\begin{tabular}{lcccc}
\hline
\textbf{Feature Type} & \textbf{Mean Effect} & \textbf{Median} & \textbf{Max} & \textbf{Neg. Effect} \\ \hline
Embeddings     & $0.01\%$             & $0.00\%$       & $0.71\%$     & $51\%$               \\
Attn. Heads    & $0.01\%$             & $0.00\%$        & $0.25\%$     & $42\%$               \\
FFN Outputs    & $0.03\%$             & $0.02\%$        & $0.38\%$     & $41\%$               \\ \hline
\end{tabular}%
}
\end{table}

Notably, roughly half of all ablations resulted in a \textit{negative} effect (increasing the stereotype likelihood), peaking at $51\%$ for Textual Embeddings. These findings heavily suggest that individual neurons are not the primary driving force of stereotypes in Large Language Models (LLMs); rather, biased representations are likely emergent from the collective activity of highly redundant circuits.

\subsection{Attention Head Results}
For GPT-2 Small, our best probe achieved approximately 73\% accuracy, while the Llama 3.2 1B probe achieved an 80\% probe accuracy.

\begin{figure*}[htbp]  
    \centering
    \begin{subfigure}{0.95\textwidth}
        \centering
        \includegraphics[width=0.8\textwidth]{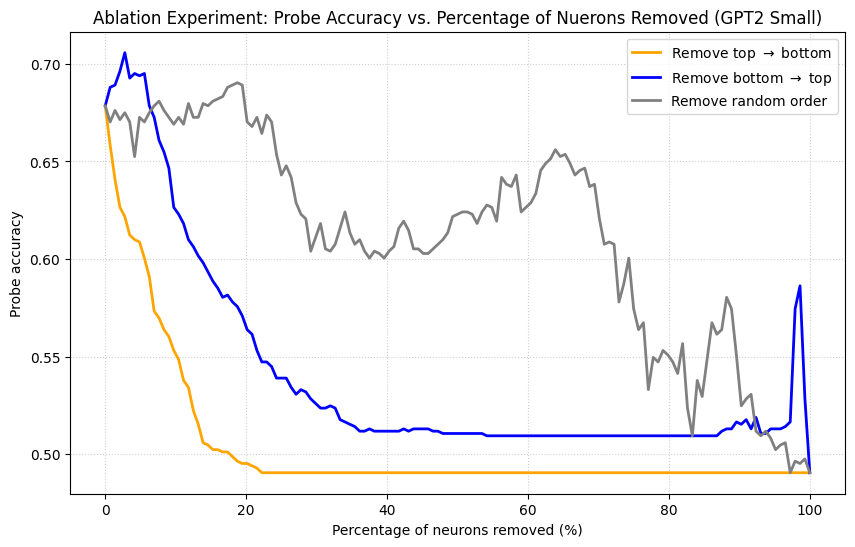}
        \caption{Attention head ablation.}
    \end{subfigure}
    \vspace{0.3em}
    \begin{subfigure}{0.95\textwidth}
        \centering
        \includegraphics[width=0.8\textwidth]{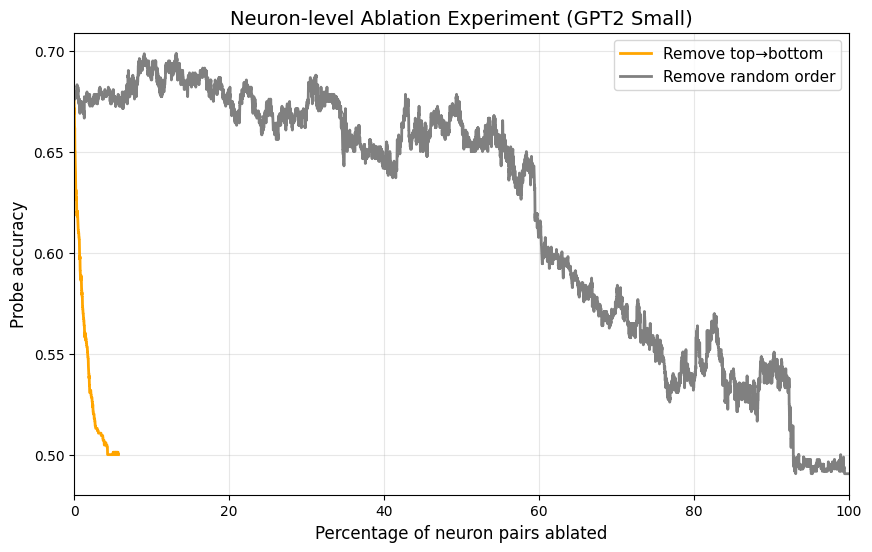}
        \caption{Neuron ablation.}
    \end{subfigure}
    \caption{Experiment 2: Probing classifier on GPT-2 Small attention head encodings. Accuracy is plotted during iterative ablation of attention heads (top) and high-impact neurons (bottom). Top-down removal follows Shapley value rankings.  {Note: For attention heads, random-guess accuracy (50\%) is reached at $\sim$20\% ablation, whereas individual neuron ablation reaches 50\% accuracy significantly faster at $\sim$5\% ablation.}}
    \label{fig:GPT2_mha_ablation_results}
\end{figure*}

\begin{figure*}[htbp]  
    \centering
    \begin{subfigure}{0.95\textwidth}
        \centering
        \includegraphics[width=0.8\textwidth]{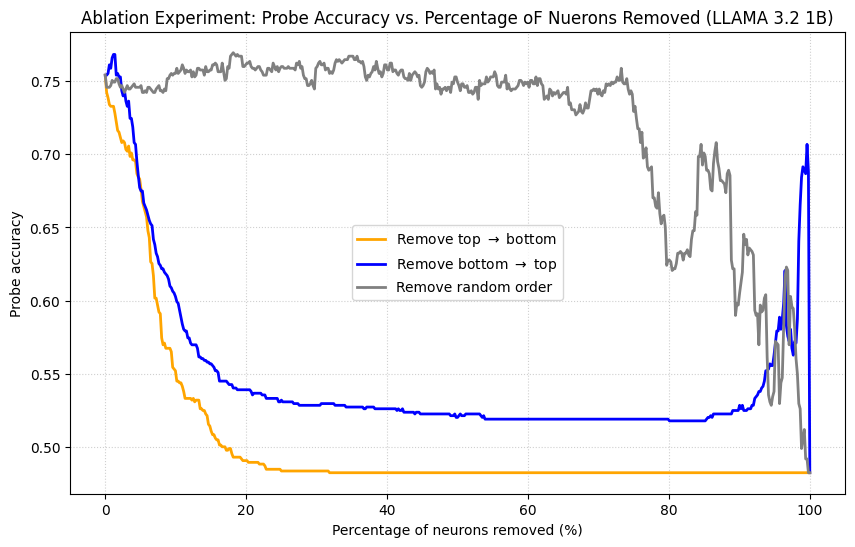}
        \caption{Attention head ablation.}
    \end{subfigure}
    \vspace{0.3em}
    \begin{subfigure}{0.95\textwidth}
        \centering
        \includegraphics[width=0.8\textwidth]{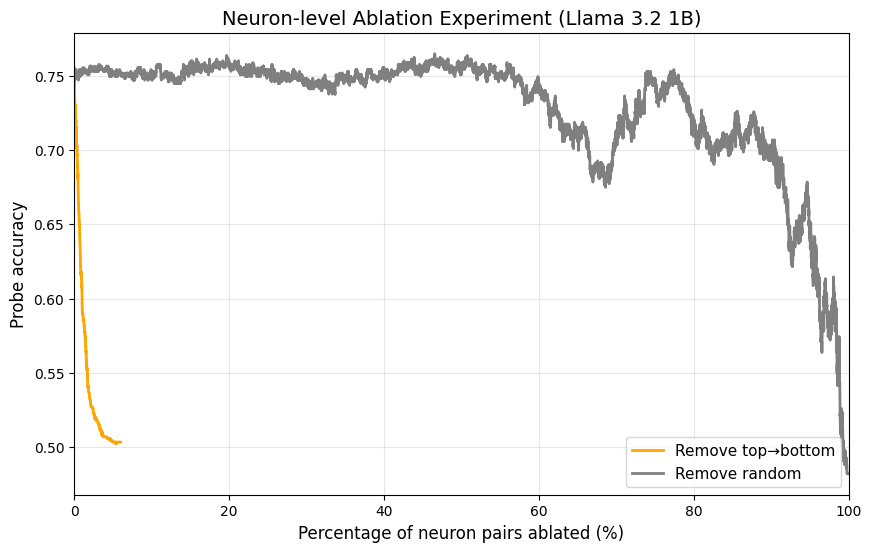}
        \caption{Neuron ablation.}
    \end{subfigure}
    \caption{Experiment 2: Probing classifier on Llama attention head encodings. Accuracy is plotted during iterative ablation of attention heads (top) and high-impact neurons (bottom). Top-down removal follows Shapley value rankings. {Note: Individual neuron ablation demonstrates a similar efficiency to the GPT-2 Small model, collapsing classifier accuracy to baseline levels ($\sim$50\%) with significantly fewer components ($\sim$5\%) compared to attention head ablation.}}
    \label{fig:Llama_mha_ablation_results}
\end{figure*}

Our results of the ablation of attention heads align with \cite{ma-etal-2023-deciphering}, which report that around 15–30\% of attention heads contribute to the model's ability to detect and encode stereotypes. Our plot shows that after around 20\% of neurons ablated when ablating full attention heads the encodings become indistinguishable. Interestingly, we can dive deeper and find a subset of neurons that encode the stereotype. As shown in for GPT-2 Small in Figure~\ref{fig:GPT2_mha_ablation_results}, around 5\% of neurons disrupt the classifier's accuracy to 50\% and as a baseline we randomly removed neuron pairs. Llama 3.2 1B also shows similar results (see Figure~\ref{fig:Llama_mha_ablation_results}).

To evaluate if this subset of neurons also contribute to the likelihood of stereotype generation, we conducted a follow-up ablation experiment targeting the top neurons with the highest Shapley values. We calculate the Stereotype Score (SS), Language Modeling Score (LMS), and Idealized Context Awareness Test (iCAT). These are scores from the \cite{Nadeem-etal-2021-stereoset} to score whether the language model is biased and if the model still functions as a coherent language model.

By zeroing out the activations of the top neurons with positive Shapley values, we compared the results.

\begin{table}[htbp]
\centering
\caption{Experiment 2: Stereoset Performance Metrics. (B) represents Baseline performance, and (A) represents the Model under Ablation of high-impact neurons identified via Shapley value analysis. In almost all iterations, the intervention successfully reduced the Stereotype Score (SS) toward the ideal 50.0 parity mark. Notably, the Language Modeling Score (LMS) remained stable or showed negligible degradation, leading to a slight net increase in the Idealized Context Awareness Test (iCAT) score for both models.}
\label{tab:stereoset_results}
\small
\begin{tabularx}{\columnwidth}{@{}l X ccc @{}}
\toprule
{Model} & {Config.} & {SS} & {LMS} & {iCAT} \\ \midrule
{Llama 3.2} & Inter (B) & 51.72 & 84.79 & 81.87 \\
(1B)               & Inter (A) & 51.11 & 84.41 & 82.54 \\ 
\cmidrule(lr){2-5}
                   & Intra (B) & 64.72 & 97.48 & 68.78 \\
                   & Intra (A) & 64.25 & 97.53 & 69.74 \\
\cmidrule(lr){2-5}
                   & {Avg (B)} & {58.22} & {91.14} & {75.33} \\
                   & {Avg (A)} & {57.68} & {90.97} & {76.14} \\ \midrule
{GPT-2}     & Inter (B) & 50.16 & 85.07 & 84.79 \\
(Small)            & Inter (A) & 50.02 & 85.07 & 85.03 \\ 
\cmidrule(lr){2-5}
                   & Intra (B) & 61.78 & 96.49 & 73.76 \\
                   & Intra (A) & 61.68 & 96.25 & 73.76 \\
\cmidrule(lr){2-5}
                   & {Avg (B)} & {55.97} & {90.78} & {79.28} \\
                   & {Avg (A)} & {55.85} & {90.66} & {79.39} \\ \midrule
\textbf{Ideal Model} & --- & \textbf{50.00} & \textbf{100.0} & \textbf{100.0} \\ \bottomrule
\addlinespace[1ex]
\multicolumn{5}{p{\columnwidth}}{
  \begin{minipage}{\columnwidth}
    \footnotesize
    {(B)} Baseline; {(A)} Ablated intervention. \\
    {SS}: Stereotype Score; \\
    {LMS}: Language Modeling Score; \\
    {iCAT}: Idealized Context Awareness Test . \\
    Note: Ablating these high impacting neurons reduces stereotype score slightly, but not significantly.
  \end{minipage}
}
\end{tabularx}
\end{table}

Causal intervention resulted in a minimal reduction in social bias across both models. Specifically, the Stereotype Score (SS) shifted toward the ideal parity of 50.0 \%, indicating a reduction in stereotypical preference in Table \ref{tab:stereoset_results}. Crucially, this mitigation was achieved without significant degradation of the Language Modeling Score (LMS), leading to a net increase in the iCAT score for both Llama-3.2-1B and GPT-2 Small.

\subsection{Embedding Results}

We also conducted study to determine whether positional and encoding embeddings independently provide sufficient signal for accurate stereotype classification. We constructed a feature space by concatenating stereotype and anti-stereotype token embeddings, with the order randomized ($p=0.5$) to prevent the classifier from learning sequence-based heuristics. To isolate the source of encoded bias, we trained separate classifiers on the encoding embeddings versus isolated positional embeddings.

As summarized in Table \ref{tab:embedding_results}, in GPT-2 Small the encoding embeddings achieved a peak validation accuracy of 0.7340, while positional embeddings performed near random chance (0.5225). This indicates that stereotypical associations are primarily localized within encoding layers rather than positional metadata.

\begin{table}[ht]
\centering
\small 
\caption{Classification Accuracy from Embeddings by Bias Type (GPT-2 Small)}
\label{tab:embedding_results}
\begin{tabular}{@{}l cc @{}} 
\toprule
\textbf{Bias Category} & \textbf{Encoding} & \textbf{Positional} \\ \midrule
Race            & 0.780          & 0.506          \\
Profession      & 0.711          & 0.528          \\
Gender          & 0.663          & 0.576          \\
Religion        & 0.634          & 0.512          \\ \midrule
\textbf{Overall Acc.} & \textbf{0.7340} & \textbf{0.5225} \\ \bottomrule
\end{tabular}
\end{table}

\section{Negative Results Explained}
While the probing analysis in Experiment 2 (Figures \ref{fig:GPT2_mha_ablation_results} and \ref{fig:Llama_mha_ablation_results}) demonstrated that a small subset of neurons is statistically necessary for a classifier to distinguish stereotypical activations, the causal intervention results in Table \ref{tab:stereoset_results} reveal a discrepancy. Ablating these ``high-impact'' neurons resulted in only marginal shifts in the StereoSet scores. This divergence between probing accuracy and causal influence can be attributed to several architectural factors inherent to the Transformer.

\label{sec:Results Explained}
\subsection{Mathematical Formulation of Information Flow}
The progression of a sequence through the GPT-2 Transformer can be formally modeled as a series of additive updates to a high-dimensional vector space known as the residual stream. Let $x_0 \in \mathbb{R}^{d_{model}}$ represent the initial input, formed by the sum of word token embeddings (WTE) and positional encodings (WPE). The state of the residual stream at any layer $L$ is defined by the accumulation of outputs from preceding sub-layers:

\begin{equation}
    x_0 = \text{Embedding}(\text{tokens}) + \text{PositionalEncoding}
\end{equation}

For each layer $i \in \{1, \dots, L\}$, the state is updated using two sub-layers with residual (skip) connections:
\begin{align}
    x_{i}^{mid} &= \text{LN}(x_{i-1}) + \text{MHA}_i(\text{LN}(x_{i-1})) \\
    x_{i} &= \text{LN}(x_{i}^{mid}) + \text{FFN}_i(\text{LN}(x_{i}^{mid}))
\end{align}

where $\text{MHA}_i$ and $\text{FFN}_i$ represent the Multi-Head Attention and Feed-Forward Network contributions at layer $i$, respectively. The final output distribution $y$ is then computed by projecting the terminal state $x_L$ back onto the vocabulary space:

\begin{equation}
y = \text{Softmax}(W_u \cdot \text{LN}(x_L))
\end{equation}

\subsection{The Pathway Hypothesis and Ablation Resistance}
Our experimental results present a paradox: while we identified 400 neurons with high Shapley values dropping a classifier's accuracy to 50\% indicating they are necessary for a classifier to differentiate stereotypical from anti-stereotypical embeddings, their ablation did not significantly reduce the model's likelihood of generating biased text. 

Mathematically, we hypothesize that stereotypical information is not localized in a group of "biased neurons" but is represented as a bias direction $v_{bias}$ within the residual stream. If this direction is supported by a large basis of neurons $\{n_1, n_2, \dots, n_k\}$, then zeroing a small subset $k' \subset k$ (the 400 ablated neurons) merely performs a partial projection:

\begin{equation}
v_{bias}' = v_{bias} - \sum_{j \in k'} \text{proj}_{n_j}(v_{bias})
\end{equation}

Because the residual stream is highly redundant, the remaining components $\sum_{j \notin k'} n_j$ continue to carry the biased signal to the final layer. Furthermore, the additive nature of the architecture creates parallel pathways. Even if a specific MHA pathway is interrupted, the "bias signal" may bypass that layer via the residual connection or be re-inserted by attention heads that attend to the initial encoding $x_0$, which we demonstrated already contains significant stereotypical signal (73\% accuracy). The model inherits bias from the encodings. 

\subsection{Empirical Example}
To make the \textit{Pathway Hypothesis} concrete, consider a minimal toy example with $d_{model} = 4$. Suppose we have identified a ``bias direction'' in the residual stream $v_{bias} = [1,\ 1,\ 0,\ 0]$, and the residual stream after the embedding layer is $x_0 = [1.2,\ 0.8,\ -0.3,\ 0.5]$.
The stereotypical signal strength is quantified by projecting $x_0$ onto $v_{bias}$:
\begin{equation}
\begin{split}
    \text{proj}_{\mathbf{v}_{bias}}(x_0) 
    &= \frac{x_0 \cdot \mathbf{v}_{bias}}{\|\mathbf{v}_{bias}\|^2} \cdot \mathbf{v}_{bias} \\
    &= \frac{1.2 + 0.8}{2} \cdot [1,1,0,0]
\end{split}
\end{equation}
yielding a scalar signal strength of $1.0$, consistent with the strong stereotypical signal already present in the token encodings (73\% probe accuracy).
\paragraph{After Layer 1.}
The MHA and FFN contributions produce an additive update written into the residual stream via the skip connection:
\begin{align}
    x_0 &= [1.2,\ 0.8,\ -0.3,\ 0.5] \\
    \Delta x_1 &= [0.1,\ -0.1,\ 0.5,\ -0.2] \\
    x_1 &= x_0 + \Delta x_1 = [1.3,\ 0.7,\ 0.2,\ 0.3]
\end{align}
The signal strength remains at $\text{proj}_{v_{bias}}(x_1) = (1.3 + 0.7)/2 = 1.0$, consistent with probe accuracy remaining stable across layers.
\paragraph{Ablating the High-Shapley Neurons.}
The 400 neurons identified by our Shapley analysis contribute a vector $\delta$ strongly aligned with $v_{bias}$:
\begin{equation}
    \delta = [0.6,\ 0.4,\ 0.0,\ 0.0]
\end{equation}
Zeroing these neurons yields the ablated residual stream:
\begin{align}
    x_1 &= [1.3,\ 0.7,\ 0.2,\ 0.3] \\ 
    \delta &= [0.6,\ 0.4,\ 0.0,\ 0.0] \\
    x_1^{\text{ablated}} &= x_1 - \delta \\
    &= [0.7,\ 0.3,\ 0.2,\ 0.3] \nonumber
\end{align}
with a new signal strength of $\text{proj}_{v_{bias}}(x_1^{\text{ablated}}) = (0.7 + 0.3)/2 = 0.5$.
Ablating the identified neurons reduces the projection onto $v_{bias}$ from $1.0$ to $0.5$, but non-negligible signal strength remains in the residual stream, consistent with the bias direction being encoded in a distributed, redundant fashion across many components of $x_L$.

\paragraph{Note on idealization.} This toy example represents $v_{bias}$ as a single linear direction in the residual stream. In practice, our classifier — a feedforward network with GELU activations — captures non-linear structure that a single vector cannot fully represent, suggesting the true bias subspace is higher-dimensional and more complex. The projection onto $v_{bias}$ is therefore a simplification for illustrative purposes; in reality our probe has likely recovered only a partial approximation of the full stereotype direction, and the residual signal strength after ablation may be larger than this idealized example suggests.

\section{Conclusion}
This study demonstrates that stereotypes in LLMs are not isolated defects but are deeply integrated into the model’s fundamental linguistic framework. Our findings are summarized as follows:

\begin{itemize}\item \textbf{Experiment 1 (Localization):} We identified individual neurons with extreme contrastive activation ratios (up to $7.1 \times 10^5$). However, single-neuron ablation proved causally insufficient, yielding a negligible $\le 0.03\%$ reduction in stereotypical likelihood.

\item \textbf{Experiment 2 (Circuit Analysis):} Probing classifiers reached $73\%$ (GPT-2) and $80\%$ (Llama 3.2) accuracy in identifying biased activations. Targeted ablation of high-impact neurons consistently improved the \textbf{iCAT} score, proving that while a "bias circuit" can be localized, its functional redundancy limits the efficacy of simple ablation.
\end{itemize}

\textbf{The Ablation Paradox:} The discrepancy between high probing accuracy and low causal impact suggests that stereotypes are represented as high-dimensional \textbf{directions} in the residual stream rather than discrete \textbf{units}. Because these signals are present in initial embeddings and bypass layers via skip connections, they are highly resistant to localized interventions.

\section{Future Work}
\label{sec:future} 

The marginal causal impact observed in Section~\ref{sec:results}---where individual neuron ablation failed to significantly reduce stereotypic likelihood---suggests that GPT-2 and Llama 3.2 represents social biases in a state of polysemanticity. In this state, a single neuron may represent multiple, unrelated concepts, a phenomenon attributed to the model's attempt to represent more features than it has available dimensions, a concept known as \textit{superposition}.

As a result, ablating a single dense neuron or a group of neurons likely removes only a small fraction of a "bias circuit" while simultaneously damaging unrelated functional circuits. To address this, future work could employ Sparse Autoencoders (SAEs) as proposed by \citet{cunningham2023sparse}.  

By training an SAE on the FFN activations, we could "untangle" the stereotype direction. In this expanded latent space, we hypothesize that stereotypical representations will occupy monosemantic latent features. Unlike the dense neurons ablated in this study, suppressing these sparse features would likely show a significantly higher causal effect on model output, providing a more surgical mechanism for bias mitigation while preserving the model's general performance.

Another avenue for future work is to develop a more generalizable classifier suitable 
for integration into production-level LLMs. Specifically, since individual layers 
achieved classification accuracy comparable to using all layers combined in miscellaneous experiments, it may be 
sufficient to extract activations from a single intermediate layer and pass them 
through a lightweight probe to determine with some confidence whether an output 
is stereotypical or biased. This approach is attractive for deployment scenarios, 
as it provides a computationally efficient mechanism for mid-inference guardrails 
without requiring full activation extraction across all layers. 

\newpage
\clearpage

\bibliographystyle{acl_natbib}
\bibliography{references}

@misc{Altman800M,
  author       = {Shepardson, Rachael},
  title        = {Sam Altman says {ChatGPT} has hit 800M weekly active users},
  howpublished = {TechCrunch},
  year         = {2025},
  month        = oct,
  day          = {6},
  url          = {https://techcrunch.com/2025/10/06/sam-altman-says-chatgpt-has-hit-800m-weekly-active-users/},
  note         = {Accessed: 2026-03-26}
}

@inproceedings{Nadeem-etal-2021-stereoset,
  title        = "{S}tereo{S}et: Measuring stereotypical bias in pretrained language models",
  author       = "Nadeem, Moin and Bethke, Anna and Reddy, Siva",
  booktitle    = "Proceedings of the 59th Annual Meeting of the Association for Computational Linguistics and the 11th International Joint Conference on Natural Language Processing (Volume 1: Long Papers)",
  month        = aug,
  year         = "2021",
  address      = "Online",
  publisher    = "Association for Computational Linguistics",
  url          = "https://aclanthology.org/2021.acl-long.416",
  pages        = "5356--5371"
}

@inproceedings{Lundberg2017Unified,
  title        = {A Unified Approach to Interpreting Model Predictions},
  author       = {Lundberg, Scott M. and Lee, Su-In},
  booktitle    = {Advances in Neural Information Processing Systems 30},
  editor       = {I. Guyon and U. V. Luxburg and S. Bengio and H. Wallach and R. Fergus and S. Vishwanathan and R. Garnett},
  pages        = {4765--4774},
  year         = {2017},
  publisher    = {Curran Associates, Inc.},
  url          = {https://proceedings.neurips.cc/paper/2017/file/8a20a8621bd5394302d7ad74e14e1fa1-Paper.pdf}
}

@inproceedings{vaswani2017attention,
  author       = {Vaswani, Ashish and Shazeer, Noam and Parmar, Niki and Uszkoreit, Jakob and Jones, Llion and Gomez, Aidan N. and Kaiser, {\L}ukasz and Polosukhin, Illia},
  title        = {Attention Is All You Need},
  booktitle    = {Advances in Neural Information Processing Systems 30},
  pages        = {5998--6008},
  year         = {2017},
  publisher    = {Curran Associates, Inc.},
  url          = {https://proceedings.neurips.cc/paper/2017/file/3f5ee243547dee91fbd053c1c4a845aa-Paper.pdf}
}

@inproceedings{heilbron2019tracking,
  author       = {Heilbron, Micha and de Lange, Floris P.},
  title        = {Tracking Naturalistic Linguistic Predictions with Deep Neural Language Models},
  booktitle    = {Proceedings of the 2019 Conference on Cognitive Computational Neuroscience},
  year         = {2019},
  month        = sep,
  doi          = {10.32470/CCN.2019.1096-0},
  url          = {https://doi.org/10.32470/CCN.2019.1096-0}
}

@inproceedings{davidson2025cxad,
  title        = {{CXAD}: Contrastive Explanations for Anomaly Detection: Algorithms, Complexity Results and Experiments},
  author       = {Davidson, Ian and Kennedy, Nicol{\'a}s and Ravi, S. S.},
  booktitle    = {Transactions on Machine Learning Research},
  year         = {2025},
  month        = jun,
  note         = {Reviewed on OpenReview},
  url          = {https://openreview.net/pdf?id=Tnwci2kLna}
}

@article{strumbelj2014explaining,
  title        = {Explaining prediction models and individual predictions with feature contributions},
  author       = {{\v{S}}trumbelj, Erik and Kononenko, Igor},
  journal      = {Knowledge and Information Systems},
  volume       = {41},
  number       = {3},
  pages        = {647--665},
  year         = {2014},
  publisher    = {Springer}
}

@inproceedings{hendrycks2016gelu,
  title        = {Gaussian Error Linear Units ({GELUs})},
  author       = {Hendrycks, Dan and Gimpel, Kevin},
  booktitle    = {arXiv preprint arXiv:1606.08415},
  year         = {2016},
  url          = {https://arxiv.org/abs/1606.08415}
}

@inproceedings{ma-etal-2023-deciphering,
  title        = "Deciphering Stereotypes in Pre-Trained Language Models",
  author       = "Ma, Weicheng and Scheible, Henry and Wang, Brian and Veeramachaneni, Goutham and Chowdhary, Pratim and Sun, Alan and Koulogeorge, Andrew and Wang, Lili and Yang, Diyi and Vosoughi, Soroush",
  booktitle    = "Proceedings of the 2023 Conference on Empirical Methods in Natural Language Processing",
  month        = dec,
  year         = "2023",
  address      = "Singapore",
  publisher    = "Association for Computational Linguistics",
  url          = "https://aclanthology.org/2023.emnlp-main.697",
  doi          = "10.18653/v1/2023.emnlp-main.697",
  pages        = "11328--11345"
}

@inproceedings{yang2025biasahead,
  title        = {Bias A-head? Analyzing Bias in Transformer-Based Language Model Attention Heads},
  author       = {Yang, Yi and Duan, Hanyu and Abbasi, Ahmed and Lalor, John P. and Tam, Kar Yan},
  booktitle    = {Proceedings of the 5th Workshop on Trustworthy NLP (TrustNLP 2025)},
  year         = {2025},
  pages        = {276--290},
  address      = {Albuquerque, New Mexico},
  publisher    = {Association for Computational Linguistics},
  doi          = {10.18653/v1/2025.trustnlp-main.18},
  url          = {https://aclanthology.org/2025.trustnlp-main.18/}
}

@inproceedings{cunningham2023sparse,
  title        = {Sparse Autoencoders Find Highly Interpretable Features in Language Models},
  author       = {Cunningham, Hoagy and Ewart, Aidan and Riggs, Logan and Huben, Robert and Sharkey, Lee},
  booktitle    = {arXiv preprint arXiv:2309.08600},
  year         = {2023},
  url          = {https://arxiv.org/abs/2309.08600},
  eprint       = {2309.08600},
  archivePrefix = {arXiv},
  primaryClass = {cs.LG}
}

\end{document}